\documentclass[10pt,twocolumn,letterpaper]{article}

\usepackage{cvpr}
\usepackage{times}
\usepackage{epsfig}
\usepackage{graphicx}
\usepackage{amsmath}
\usepackage{bm}
\usepackage{amssymb}
\usepackage{enumerate}
\usepackage{multirow}
\usepackage{array}

\usepackage[pagebackref=true,breaklinks=true,letterpaper=true,colorlinks,bookmarks=false]{hyperref}

\cvprfinalcopy 
\DeclareMathOperator*{\argmax}{arg\,max} 
\DeclareMathOperator*{\argmin}{arg\,min} 

\def\cvprPaperID{****} 

\begin{document}

\title{Towards Interpretable Deep Neural Networks by Leveraging \\ Adversarial Examples}

\author{Yinpeng Dong \hspace{1.2cm} Hang Su \hspace{1.2cm} Jun Zhu \hspace{1.2cm} Fan Bao \\
Tsinghua National Lab for Information Science and Technology\\
 State Key Lab of Intelligent Technology and Systems\\
Center for Bio-Inspired Computing Research\\
Department of Computer Science and Technology, Tsinghua University\\
\tt\small{\{dyp17@mails,suhangss@mail,dcszj@mail,baof14@mails\}.tsinghua.edu.cn}
}

\maketitle

\begin{abstract}
   Deep neural networks (DNNs) have demonstrated impressive performance on a wide array of tasks, but they are usually considered opaque since internal structure and learned parameters are not interpretable. In this paper, we re-examine the internal representations of DNNs using adversarial images, which are generated by an ensemble-optimization algorithm. We find that: (1) the neurons in DNNs do not truly detect semantic objects/parts, but respond to objects/parts only as recurrent discriminative patches; (2) deep visual representations are not robust distributed codes of visual concepts because the representations of adversarial images are largely not consistent with those of real images, although they have similar visual appearance, both of which are different from previous findings. To further improve the interpretability of DNNs, we propose an adversarial training scheme with a consistent loss such that the neurons are endowed with human-interpretable concepts. The induced interpretable representations enable us to trace eventual outcomes back to influential neurons. Therefore, human users can know how the models make predictions, as well as when and why they make errors.
\end{abstract}

\section{Introduction}


Deep Neural Networks (DNNs) have demonstrated unprecedented performance improvements in numerous applications~\cite{DeepReview_Lecun_2015}, including speech recognition~\cite{mohamed2012acoustic,seide2011conversational}, image classification~\cite{krizhevsky2012imagenet,simonyan2014very,szegedy2015going,he2015deep}, object detection~\cite{girshick2014rich,ren2015faster}, \textit{etc}.
Nevertheless, the DNNs are still treated as ``black-box'' models due to the lack of understandable decoupled components and the unclear working mechanism~\cite{bengio2013representation}.
In some cases, the effectiveness of DNNs is limited when the models are incapable to explain the reasons behind the decisions or actions to human users, since it is far from enough to provide eventual outcomes to the users especially in mission-critical applications, \textit{e.g.}, healthcare or autonomous driving.
The users may also need to understand the rationale of the decisions such that they can understand, validate, edit, trust a learned model, and fix the potential problems when it fails or makes errors. Therefore, it is imperative to develop algorithms to learn features with good interpretability, such that the users can clearly understand, appropriately trust, and effectively interact with the models.



Many attempts have been made to address the lack of interpretability issue in DNNs.
For example, in~\cite{hendricks2016generating,park2016attentive}, a justification is provided along with a decision, to point out the visual evidence by a natural sentence.
In~\cite{zhou2016learning,selvaraju2016grad}, the predictions of a model are explained by highlighted image regions.
High-level semantic attributes can be integrated into DNNs to improve the interpretability explicitly during the learning process~\cite{dong2017improving}.
Moreover, a great amount of efforts have been devoted to visualize and interpret the internal representations of DNNs~\cite{erhan2009visualizing,simonyan2013deep,zeiler2014visualizing,zhou2014object,springenberg2014striving,liu2016towards}.
A neuron in convolutional layers can be interpreted as an object/part detector by the activation maximization~\cite{erhan2009visualizing,zhou2014object} or various gradient-based algorithms~\cite{zeiler2014visualizing,simonyan2013deep,springenberg2014striving}.

However, these methods are performed on specific datasets (\textit{e.g.}, ImageNet~\cite{russakovsky2015imagenet}, Place~\cite{zhou2014learning}), which are not comprehensively justified in the complex real world.
Additionally, they are generally conducted to interpret correct predictions, while few attentions have been paid to investigate why an error has been made. A fault confessed is half redressed. If we do not know the weaknesses of the model, we cannot fix its potential problems.


\subsection{Our Contributions}



\begin{figure*}[t]
  \centering
  \includegraphics[width=0.98\linewidth]{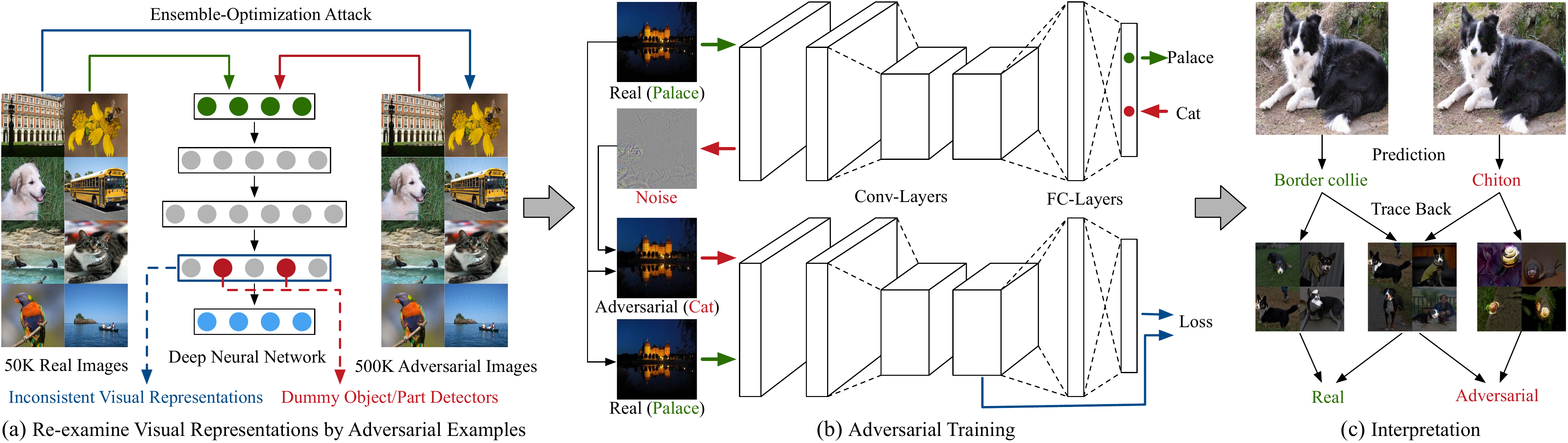}
  \caption{The overall framework in this paper. (a) We first generate a set of $500$K adversarial images by the ensemble-optimization attack and re-examine the visual representations with the conclusions of dummy object/part detectors and inconsistent visual representations. (b) We show that adversarial training facilitates the improvement of the interpretability and consistency of representations in DNNs. (c) The interpretable representations enable us to trace the eventual predictions back to influential neurons, and explain how the predictions have been made as well as when and why an error occurs.}
  \label{fig:framework}
\end{figure*}

There are many aspects of what make a model interpretable to humans. In this paper, we focus on analyzing the behavior of DNNs when facing irregular examples (\ie, adversarial examples) and explaining the predictions to human users by tracing outputs back to features. 
Specifically, we re-examine previous conclusions about interpretable representations and improve the interpretability of DNNs by leveraging adversarial examples.
With the maliciously generated adversarial examples~\cite{szegedy2013intriguing,goodfellow2014explaining}, the DNNs output attacker-desired (but incorrect) predictions with high confidence.
Adversarial examples along with the real examples enable us to investigate the behavior of DNNs from both positive and negative sides, \ie, we can analyze how DNNs make accurate predictions as well as errors, uncovering the mechanism of DNNs to some extent.
We adopt adversarial images to analyze how DNNs make errors instead of the false predictions of real images because the errors of real images are relatively negligible, \eg, the misclassification of \textit{tabby cat} to \textit{tiger cat} is more visually and semantically acceptable than to \textit{school bus}.
Based on the interpretable DNNs, we then make several steps towards the interpretability by explaining how the DNNs make decisions, as well as when and why DNNs make mistakes.
We summarize our contributions as follows.

\textbf{Adversarial dataset.}
In order to fully ascertain the behavior of DNNs, we construct a set of adversarial images.
We generate $10$ images with different target labels for each image in the ILSVRC 2012~\cite{russakovsky2015imagenet} validation set, resulting in an adversarial validation set of size $500$K.
An ensemble-optimization attack algorithm is adopted to generate more general adversarial images, which are less sensitive to a specific model and have good transferability~\cite{liu2016delving}, as is demonstrated in Fig.~\ref{fig:framework}~(a).

\textbf{Dummy object/part detectors and inconsistent visual representations.}
We re-examine the internal representations of DNNs by adversarial examples.
In the set of our experiments, we test several standard architectures including AlexNet~\cite{krizhevsky2012imagenet}, VGG~\cite{simonyan2014very} and ResNet~\cite{he2015deep} using both the real and the generated adversarial images.
We inspect the response of neurons in DNNs manually and also quantitatively evaluate the alignment of visual concepts in selected real and adversarial images, which activate a given neuron.
Interestingly, we find that: (1) the neurons with high-level\footnote{In this paper, we mean the high-level features by objects and parts, while low-level features include colors and textures, and mid-level features include attributes (\eg, shiny) and shapes.} semantic meanings found by real images do not reveal the same patterns when showing adversarial images.
We conclude that neurons in DNNs do not truly detect semantic objects/parts, but respond to objects/parts only as recurrent discriminative patches, which is contradictory to previous findings~\cite{zhou2014object,bau2017network,gonzalez2016semantic}; (2) we also demonstrate that the overall representations of adversarial images are inconsistent with those of the corresponding real images, so the deep visual representations are not robust distributed codes of visual concepts, contrary to~\cite{agrawal2014analyzing}, as is shown in Fig.~\ref{fig:framework}~(a).

\textbf{Improve the interpretability of DNNs with adversarial training.}
Adversarial training has proved to be an effective way to improve the robustness of DNNs~\cite{goodfellow2014explaining,kurakin2016adversarial}.
In this paper, we propose to improve the interpretability of internal representations by designing an appropriate adversarial training scheme.
By recovering from the adversarial noise, the representation of an adversarial image resembles that of the corresponding real image.
The procedure encourages the neurons to learn to resist the interference of adversarial perturbations, and thus the neurons are consistently activated when the preferred objects/parts appear, while deactivated when they disappear, as shown in Fig.~\ref{fig:framework}~(b).
The interpretable representations provide a cue to explain the rationale of the model's predictions, \ie, human users can trace a prediction back to particularly influential neurons in the decision making procedure.
Furthermore, human users can also know when and why the model makes an error by leveraging the interpretable representations, \ie, which group of neurons contribute to such a flaw prediction, as shown in Fig.~\ref{fig:framework}~(c).

\section{Re-examine the Internal Representations}
It is a popular statement that deep visual representations have the good transferability since they learn disentangled representations, \ie, some neurons can detect semantic objects/parts spontaneously, and thus form the human-interpretable representations~\cite{zhou2014object,zeiler2014visualizing,bau2017network}.
In this section, we criticize this traditional view by showing that the object/part detectors emerged in DNNs may be easily ``fooled'' by the irregular examples (\ie, adversarial examples) in the complex world.
There are two possible hypotheses to answer the following questions ``what are the representations of adversarial examples, and why do the representations lead to inaccurate predictions?''
\begin{itemize}
\item The representations of adversarial examples \emph{align well with the semantic objects/parts}, but are not discriminative enough, resulting in erroneous predictions.
\item The representations of adversarial examples \emph{do not align with the semantic objects/parts}, which means by adding small imperceptible noises, the neurons cannot detect corresponding objects/parts in adversarial images, leading to inaccurate predictions.
\end{itemize}
To examine the ``authentic'' internal representations of DNNs when facing adversarial examples, we conduct experiments on the ImageNet dataset~\cite{russakovsky2015imagenet} for illustration.
We first construct a set of adversarial images by the ensemble-optimization attack, and then feed them to the models to figure out the potential problems.

\subsection{Ensemble-Optimization Attack}

DNNs are shown to be vulnerable against adversarial perturbations of the input~\cite{nguyen2015deep}.
Several methods have been proposed to generate adversarial examples such as box-constrained L-BFGS~\cite{szegedy2013intriguing}, Fast Gradient Sign~\cite{goodfellow2014explaining}, DeepFool~\cite{moosavi2016deepfool}, \textit{etc}.
Since the adversarial examples generated for a specific model are model-sensitive, we instead generate them by an ensemble-optimization attack algorithm following~\cite{liu2016delving}.
The generated adversarial images are more general because different models make the same mistakes, which are more likely to be misclassified by other unseen models due to the transferability~\cite{liu2016delving}.

Suppose $\bm{x}$ is a real image with ground-truth class $y$, and $f_\theta(\bm{x})$ is a classifier. We seek to generate an adversarial image $\bm{x}^*$, which adds small imperceptible noise to $\bm{x}$ but the classifier will classify $\bm{x}^*$ to a target class as $f_\theta(\bm{x}^*)=y^*$ with $y^* \neq y$.
To fulfill the goal of attack, we solve the ensemble-optimization problem as
\begin{equation}
\argmin_{\bm{x}^*} \lambda \cdot d(\bm{x},\bm{x}^*) + \sum_{i=1}^k \ell\big(\mathbf{1}_{y^*}, f_{\theta}^{(i)}(\bm{x}^*)\big),
\label{eq:attack}
\end{equation}
where $d$ is the $L_2$ distance to quantify the difference between the real image and its adversarial counterpart, which guarantees the adversarial samples should be close to the real ones; $\mathbf{1}_{y^*}$ is the one-hot encoding of the target class $y^*$; $f_{\theta}^{(i)}$ is the output of the $i$-th model; $\ell$ is the cross entropy loss to measure the distance between the prediction and the target; $k$ is the number of ensemble models that are taken into consideration; and $\lambda$ is a constant to balance constraints. We therefore generate an adversarial image belonging to the target class $y^*$ when the optimization problem reaches its minimum, with its original class denoted as $y$.

We attack the ensemble of AlexNet~\cite{krizhevsky2012imagenet}, VGG-16~\cite{simonyan2014very} and ResNet-18~\cite{he2015deep} models.
For solving the optimization problem in Eq.~\eqref{eq:attack}, we adopt Adam optimizer with step size $5$ for $10\sim20$ iterations.
We generate $10$ adversarial images for each image in the ILSVRC 2012 validation set, with the $10$ least likely classes (least average probability), \ie, we choose 10 different $y^*$, as targets. So we construct an adversarial validation set of $500$K images (See Fig.~\ref{fig:framework}~(a) for examples).

\subsection{Dummy Object/Part Detectors}
We first show what a neuron in DNNs truly learns to detect. For each neuron, we find the top $1\%$ images with highest activations in the real and adversarial validation sets respectively (\ie, $500$ real images and $5000$ adversarial images) to represent the learned features of that neuron~\cite{girshick2014rich,zhou2014object}, making the neuron speak for itself.

\begin{figure*}
  \centering
  \includegraphics[width=0.98\linewidth]{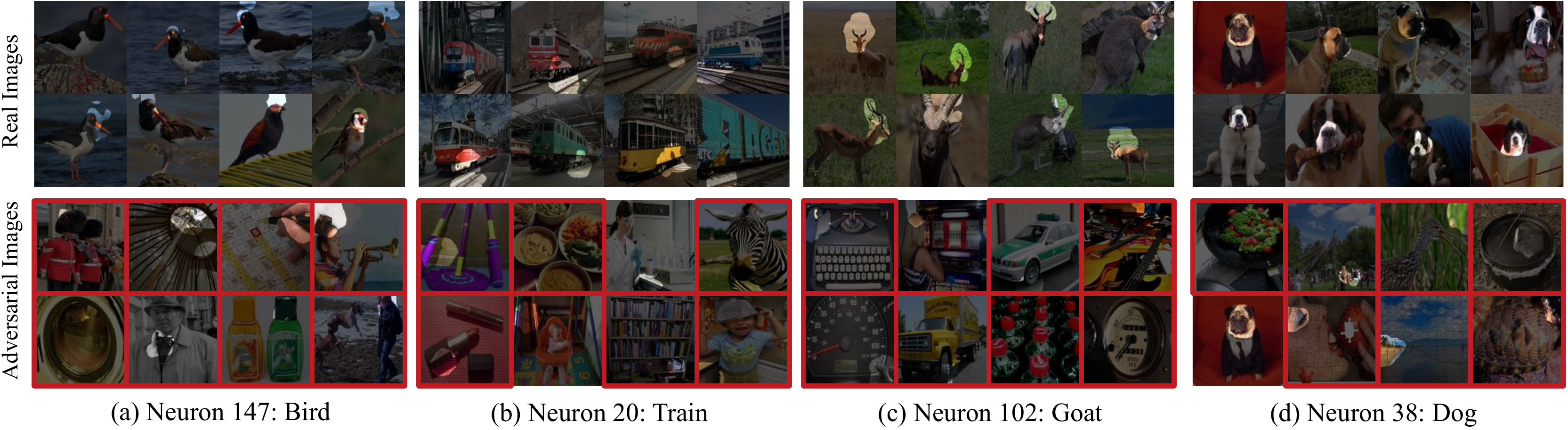}
  \caption{The real and adversarial images with highest activations for neurons in VGG-16 \textit{pool5} layer. The neurons have explicit semantic meanings in real images, which do not appear in adversarial images. The adversarial images in red boxes have the target classes the same as the meanings of the neurons (\eg, the model misclassifies the adversarial images in (a) as \textit{birds}). The highlighted regions are found by discrepancy map~\cite{zhou2014object}. More visualization results of AlexNet and ResNet-18 can be found in Appendix.}
  \label{fig:VGG16-features}
\end{figure*}

We show some visualization results in Fig.~\ref{fig:VGG16-features}, in which the highlighted regions are found by discrepancy map~\cite{zhou2014object}, \ie, the given patch
is important if there is a large discrepancy and vice versa. 
As shown in the first row of Fig.~\ref{fig:VGG16-features}, the neurons have explicit explanatory semantic meanings or human-interpretable concepts when showing real images only, but the contents of the adversarial images do not align with the semantic meanings of the corresponding neurons, if we look at the second row.
In general, the neurons tend to detect nothing in common for adversarial samples.

After investigating the behavior of neurons, we find that the neurons with high-level semantic meanings with respect to the real images are also highly activated by the adversarial images (in red boxes) of specific target classes, which are similar to the corresponding real images. Nevertheless, the visual appearance is of significant difference between the real and adversarial images.
On the other hand, the adversarial images with contents similar to the meaning of the neurons have lower activations, indicating the neurons cannot detect corresponding objects/parts in adversarial images.
For example, the \textit{neuron 147} in Fig.~\ref{fig:VGG16-features}~(a) detects the concept \textit{bird head} in real images, but it fires for adversarial images with various objects, most of which (in red boxes) are misclassified as \textit{bird} due to the adversarial attack.
Moreover, it does not fire for adversarial images with real \textit{birds}, meaning the neurons do not truly detect \textit{bird head} in input images.
So we argue that \emph{the neurons in DNNs with high-level concepts do not actually detect semantic objects/parts in input images, but they only tend to respond to discriminative patches with the corresponding output classes}.


\subsubsection{Quantification of Features}
\label{sec:quantify}

To further verify our argument, we quantitatively analyze the behavior of neurons by first proposing a metric to quantify the level and the consistency of the features learned by each neuron, and then measuring the similarity of visual concepts in real and adversarial images.
Considering of the intuitive observations, low-level concepts (such as colors, textures, \textit{etc}) can occur in a broad range of images with diverse classes, while high-level concepts (such as objects, parts) only appear in several specific classes.
It is easy to conclude that a neuron tends to consistently detect high-level semantic concepts (\ie, objects/parts) if its response concentrates on some specific classes, and otherwise spreads across a wide range of classes. Due to these observations, we propose a metric to measure the level and the consistency of the features learned by each neuron.

\begin{figure}
  \centering
    \includegraphics[width=0.98\linewidth]{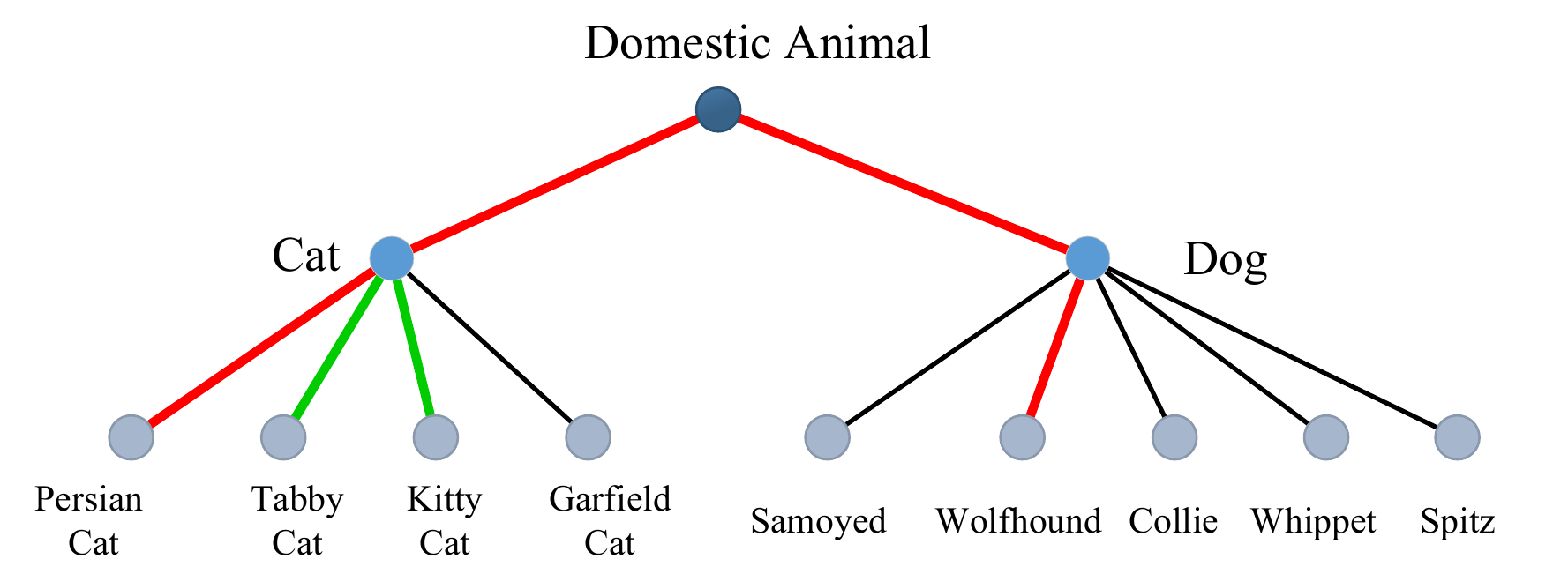}
    \caption{Illustration for quantifying the level and the consistency of features on WordNet. The red path indicates the distance between \textit{persian cat} and \textit{wolfhound} $d=4$, which is larger than the distance between \textit{tabby cat} and \textit{kitty cat} ($d=2$ indicated by the green path).}
    \label{fig:WordNet}
\end{figure}

Given a set of real images with highest activations (\ie, top $1\%$) for a neuron $n_i$, we extract a categorical distribution $\bm{p_i}$ to indicate which class it prefers, where $p_i^j$ is the fraction of images with ground-truth label $j$.
It is noted that entropy of $\bm{p_i}$, which is a natural choice to be the metric, fails to explore the hierarchical correlation between classes.
As an example in Fig.~\ref{fig:WordNet}, a neuron that responds to different types of \textit{dogs} detects a higher level of human-interpretable concepts (\eg, \textit{dog face}) than another neuron which responds to both \textit{cats} and \textit{dogs} (\eg, \textit{animal fur}).
To address this issue, we quantify the features learned by each neuron based on WordNet~\cite{miller1990introduction}, which groups different classes by means of their conceptual-semantic and lexical relations.

As is demonstrated in Fig.~\ref{fig:WordNet}, we measure the distance between different classes on the WordNet tree, \textit{e.g.}, the green path indicates $d=2$ for \textit{tabby cat} and \textit{kitty cat}, and the red path implies a large distance $d=4$ between \textit{persian cat} and \textit{wolfhound}. As the distance in the WordNet tree is an appropriate measurement for semantic similarity between words~\cite{jurafsky2014speech}, we thus define the correlation between the corresponding classes as
\begin{equation}
c_{i,j} = \exp\left(-\frac{d^2(w_i, w_j)}{2\sigma^2}\right),
\end{equation}
where $w_i$, $w_j$ are the words of the $i, j$-th classes, $d(w_i, w_j)$ is their WordNet tree distance, and $\sigma$ is a hyper-parameter to control the decaying rate, which is set to $1$ in experiments.
We form the distance matrix by collecting each pair of the corresponding classes as $\mathbf{C}=[c_{i,j}]$. The level and the consistency of the features learned by the neuron $n_i$ is quantified as the semantic distance
\begin{equation}\label{equ_INPscore}
\mathrm{LC}(n_i) = \|\bm{p_i}\|_{\mathbf{C}}^2 = \bm{p_i}^T\mathbf{C}\bm{p_i}.
\end{equation}
A higher score implies that the neuron consistently detects a higher level semantic concept, or concentrates on more specific categories.
Note that we do not specify the true meanings for the neurons like~\cite{bau2017network} but only measure the level and the consistency of their features, to fulfill the studies in the following sections.
The proposed metric correlates well with human judgment, which is discussed in Appendix.

\subsubsection{Quantitative Results}

In this section, we investigate the alignment of the detected concepts by considering the real and adversarial images with highest activations for a neuron (\ie, top $1\%$).
Specifically, we denote the categorical distribution corresponding to a given neuron as $\bm{p}$ for the top $1\%$ real images; furthermore, we have two categorical distributions for the top $1\%$ adversarial images as $\bm{q}$ and $\tilde{\bm{q}}$, which are the distributions for their original classes and target classes.

Similar as the $\mathrm{LC}$ metric in Eq.~\eqref{equ_INPscore}, we calculate the cosine similarity between the real and adversarial images as
\begin{equation}
\mathrm{CS}_1= \frac{\langle\bm{p}, \bm{q}\rangle_{\mathbf{C}}}{\|\bm{p}\|_{\mathbf{C}} \cdot \|\bm{q}\|_{\mathbf{C}}} = \frac{\bm{p}^T\mathbf{C}\bm{q}}{\sqrt{\bm{p}^T\mathbf{C}\bm{p}} \sqrt{\bm{q}^T\mathbf{C}\bm{q}}}, 
\end{equation}
\begin{equation}
\mathrm{CS}_2= \frac{\langle\bm{p}, \tilde{\bm{q}} \rangle_{\mathbf{C}}}{\|\bm{p}\|_{\mathbf{C}} \cdot \|\tilde{\bm{q}}\|_{\mathbf{C}}} =\frac{\bm{p}^T\mathbf{C}\tilde{\bm{q}}}{\sqrt{\bm{p}^T\mathbf{C}\bm{p}} \sqrt{\tilde{\bm{q}}^T\mathbf{C}\tilde{\bm{q}}}},
\end{equation}
where $\mathrm{CS}_1$ measures the similarity of contents between the real and adversarial images for a neuron; and $\mathrm{CS}_2$ is corresponding to the similarity of the predicted classes between them. As an extreme case, $\mathrm{CS}_1 = 1$ means that the original classes of adversarial images are the same as the classes of real images, meaning the same contents.
On the other hand, $\mathrm{CS}_2 = 1$ indicates that the target classes of adversarial images are the same as the real ones without regarding the true contents.
Note that $\mathrm{CS}_1$ and $\mathrm{CS}_2$ cannot be high at the same time, because $\bm{q}$ and $\tilde{\bm{q}}$ are of great difference due to the attack.

\begin{figure}[!thb]
  \centering
  \includegraphics[width=1.0\linewidth]{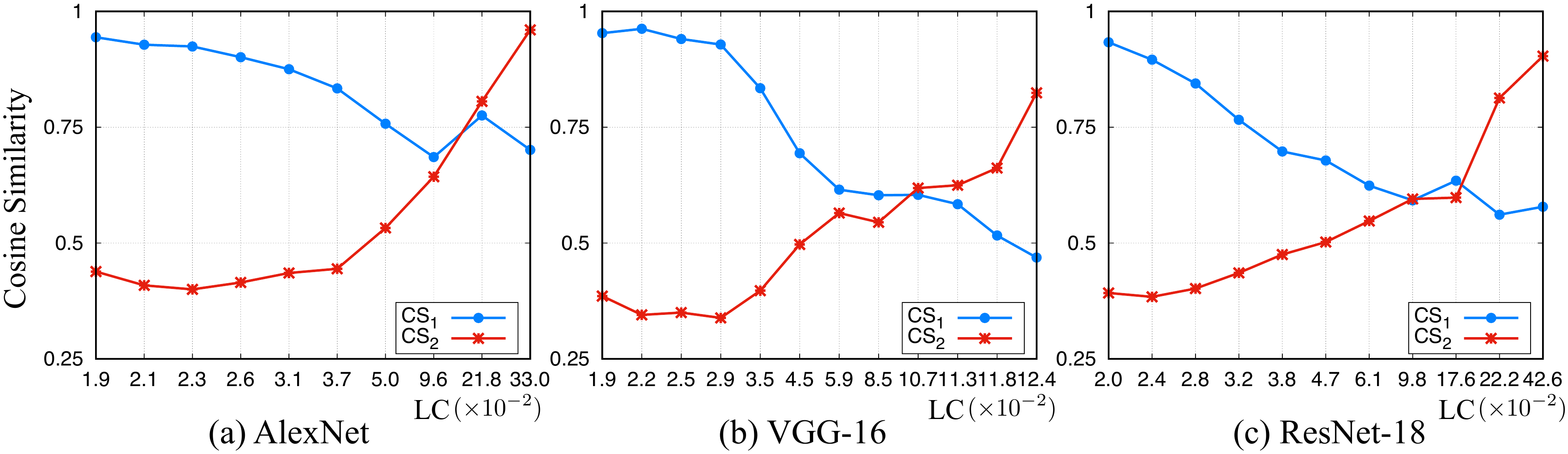}
  \caption{Cosine similarity between the classes of real and adversarial images for neurons against the level and the consistency of their features. We average the $\mathrm{CS}_1$ and $\mathrm{CS}_2$ of different neurons around a given $\mathrm{LC}$ value. The neurons come from all convolutional layers in each model.}
  \label{fig:cos-raw}
\end{figure}

The results of the cosine similarity against each neuron's $\mathrm{LC}$ score (derived by Eq.~\eqref{equ_INPscore} with the categorical distribution $\bm{p}$ of real images) is shown in Fig.~\ref{fig:cos-raw}.
With the neurons' $\mathrm{LC}$ score increasing, $\mathrm{CS}_1$ decreases which means the neurons with respect to high-level semantic concepts do not respond to the true contents in images; nevertheless, $\mathrm{CS}_2$ increases which implies that the neurons respond to similar classes although the visual inputs are not relevant. This observation indicates that the neurons with higher level meanings tend to respond to images of specific classes but not the true contents of images. It supports our argument that the neurons with high-level semantic meanings do not detect objects/parts in input images, but they only respond to discriminative patches with respect to the preferred outputs, which agrees with the second hypothesis at the beginning of this section.

\subsection{Inconsistent Deep Visual Representations}
We further analyze the behavior of the overall visual representations.
For each adversarial image $\bm{x}^*$ with its original class $y$ and target class $y^*$, we first retrieve the real images in the ILSVRC 2012 validation set of class $y$ and $y^*$ as $\{\bm{x}\}_y$ and $\{\bm{x}\}_{y^*}$, respectively.
We then calculate the ratios of the average Euclidean distance between the representations as
\begin{equation}
r_1 = \frac{\bar{d}(\phi(\bm{x}^*), \{\phi(\bm{x})\}_y)}{\bar{d}(\{\phi(\bm{x})\}_y,\{\phi(\bm{x})\}_y)},
\end{equation}
\begin{equation}
r_2 = \frac{\bar{d}(\phi(\bm{x}^*), \{\phi(\bm{x})\}_{y^*})}{\bar{d}(\{\phi(\bm{x})\}_y,\{\phi(\bm{x})\}_{y^*})},
\end{equation}
where $\bar{d}$ measures the average pairwise Euclidean distance of points and $\phi$ is the feature representation of an image in the last convolutional layer of a specific model.
$r_1$ indicates whether the representation of an adversarial image is away from the real image representations of its original class and $r_2$ measures how much is the representation close to those of its target class. The denominators in $r_1$ and $r_2$ act as normalization terms to make a consistent representation have mean $1$.

\begin{figure}[!thb]
  \centering
  \includegraphics[width=0.98\linewidth]{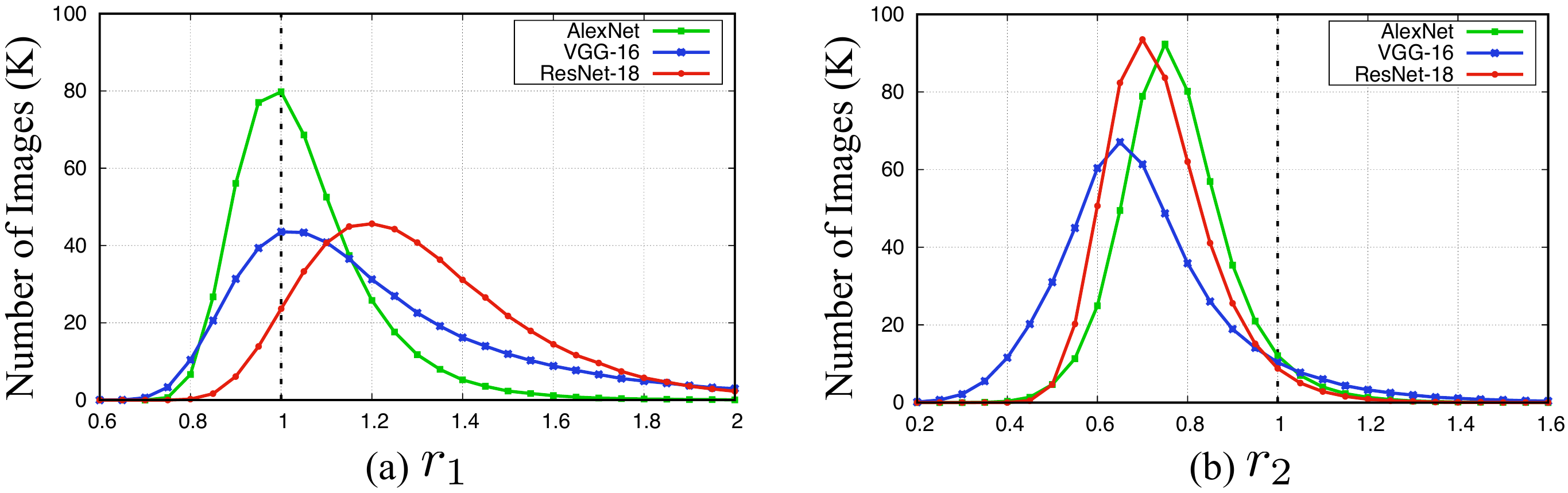}
  \caption{The distribution of adversarial images against the ratio of distance. The vertical axis shows the number of adversarial images around a given ratio. The average $r_1$ of AlexNet, VGG-16 and ResNet-18 are $1.07$, $1.26$ and $1.34$ for all adversarial images while the average $r_2$ of them are $0.80$, $0.73$ and $0.76$, respectively.}
  \label{fig:distance}
\end{figure}

We show the number of adversarial images around a given ratio in Fig.~\ref{fig:distance}.
A large number of adversarial images have $r_1$ lager than $1$, which proves that the representation of an adversarial image is far from the representations of the real images with its original class.
Moreover, great majority of adversarial images have much smaller $r_2$ than $1$, indicating that the representation of an adversarial image is closer to those of its target class than the average distance between these two classes.

From the results, we conclude that the overall visual representations of adversarial images are inconsistent with those of real images.
That means, the representations are greatly affected by adversarial perturbations and they are not robust distributed codes of visual concepts.
One consequence of this fact is that the representations will also lead to inaccurate predictions for other transferred tasks like object detection, visual question answering, video processing, \textit{etc}.
So the effectiveness of visual representations is limited by their inconsistency between adversarial images and real images.

\subsection{When Do DNNs Make Errors?}

The inconsistent feature representations provide us an opportunity to tell when the DNNs make errors.
The representations of adversarial images do not resemble those of real images, which can be detected as outliers.
We estimate the probability distribution of visual representations by a conditional Gaussian distribution as $p(\phi(\bm{x}) | y=i) = \mathcal{N}(\bm{\mu}_i, \bm{\Sigma}_i)$, where $y$ is the ground-truth label of $\bm{x}$ and $i = 1, ..., 1000$ with $1000$ being the number of classes.
We learn $\{\bm{\mu}_i, \bm{\Sigma}_i\}$ on the ILSVRC 2012 training set with images of class $i$ and estimate the log-probability of representations on real and adversarial validation sets conditioned on the predictions $\hat{y}$, where $\hat{y} = \argmax f_{\theta}(\bm{x})$ with $f_{\theta}(\bm{x})$ being the prediction of a DNN.
\begin{figure}[h]
  \centering
    \includegraphics[width=0.5\linewidth]{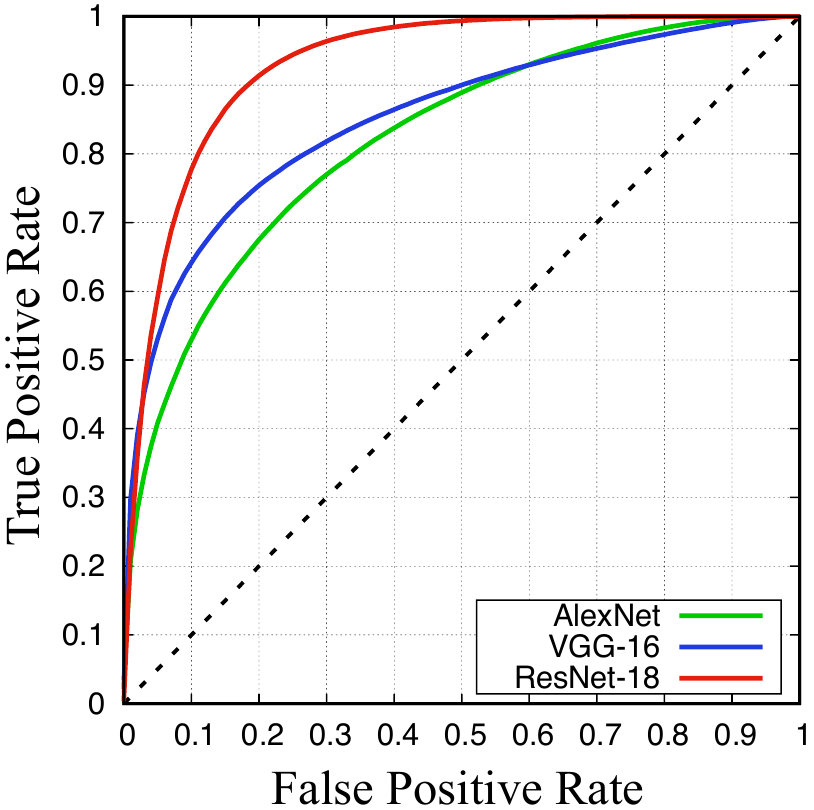}
    \caption{ROC curve for adversarial perturbation detection.}
    \label{fig:ROC}
\end{figure}

We show the detection ROC curve in Fig.~\ref{fig:ROC}. The AUC is $0.817$, $0.847$ and $0.926$ for AlexNet, VGG-16 and ResNet-18, respectively.
It shows that we can effectively detect adversarial images due to the inconsistency between the representations of real and adversarial images.
This procedure enables us to tell when the models make mistakes, but it is insufficient to tell why they make mistakes. After improving the interpretability of DNNs by adversarial training, we can answer when and why DNNs make mistakes in Sec.~\ref{sec:visual}.

\section{Adversarial Training}
The neurons that act as dummy object/part detectors and the visual representations that are inconsistent between real and adversarial images make the interpretability unreliable in DNNs.
In this section, we propose to improve the interpretability of DNNs with adversarial training.
Unlike other methods that use high-level semantic attributes to improve the interpretability of DNNs explicitly~\cite{dong2017improving}, we train the DNNs towards interpretability implicitly.
Adversarial training has the potential to train interpretable DNNs because it makes the models learn more robust concepts in input space, yielding the representations of adversarial images resemble those of the original images by suppressing the perturbations.
To achieve that, we introduce a consistent (feature matching) loss in adversarial training because that the supervision in output can only make the input-output mapping smoother, while not guarantee the input-representation mapping smoother.
The consistent loss term aims to recover from the adversarial noises in representations, which makes the neurons consistently activate when the preferred objects/parts appear without the interference of adversarial noises.

\begin{figure*}[t]
  \centering
  \includegraphics[width=0.98\linewidth]{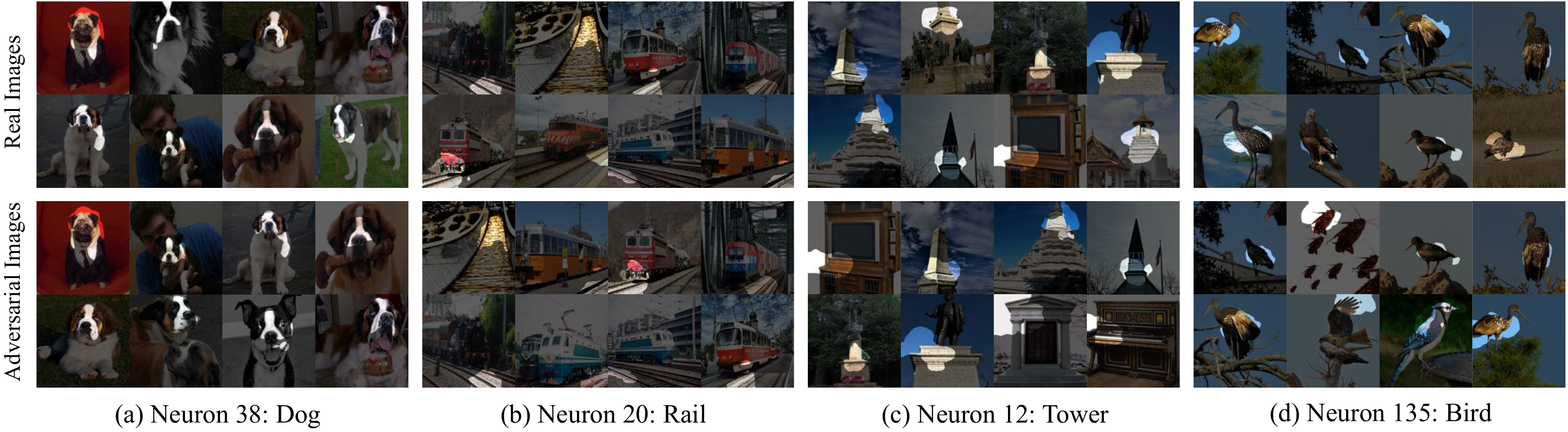}
  \caption{The real and adversarial images with highest activations for neurons in VGG-16 \textit{pool5} layer after adversarial training. The visual concepts in both of them are quite similar. More visualization results of AlexNet and ResNet-18 are provided in Appendix.}
  \label{fig:VGG16-features-adv}
\end{figure*}

\begin{figure}[t]
  \centering
  \includegraphics[width=1.0\linewidth]{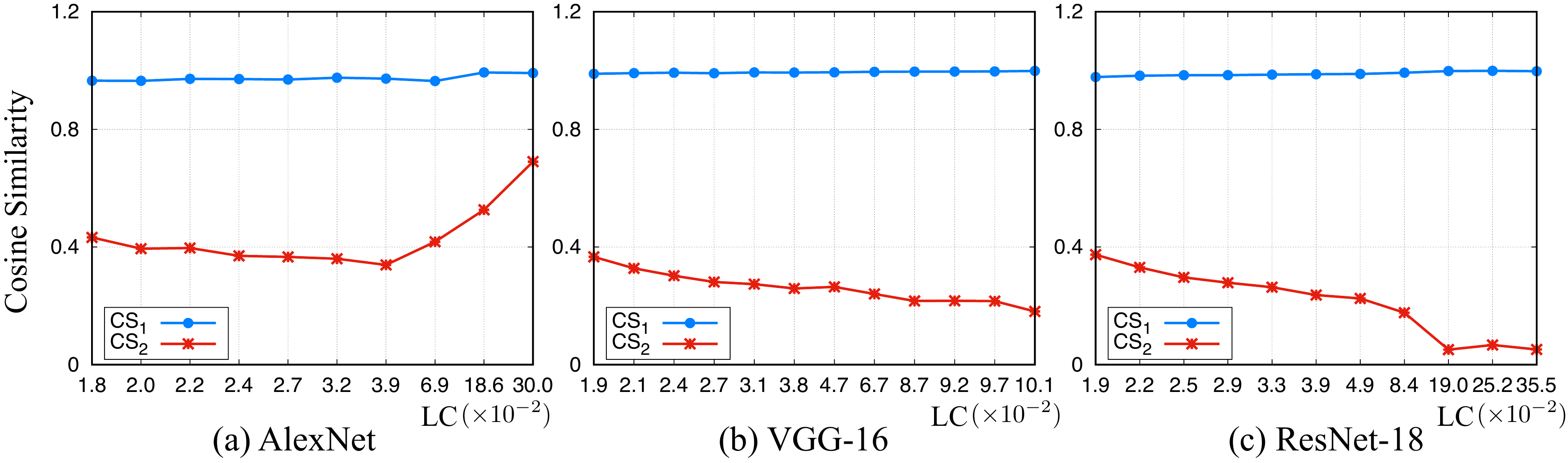}
  \caption{Cosine similarity between the classes of real and adversarial images for neurons against the level and the consistency of their features after adversarial training. $\mathrm{CS}_1$ remains high and $\mathrm{CS}_2$ remains low when the $\mathrm{LC}$ score increases, which proves the neurons learn more robust concepts.}
  \label{fig:cos-train}
\end{figure}

\begin{figure}[t]
  \centering
  \includegraphics[width=0.98\linewidth]{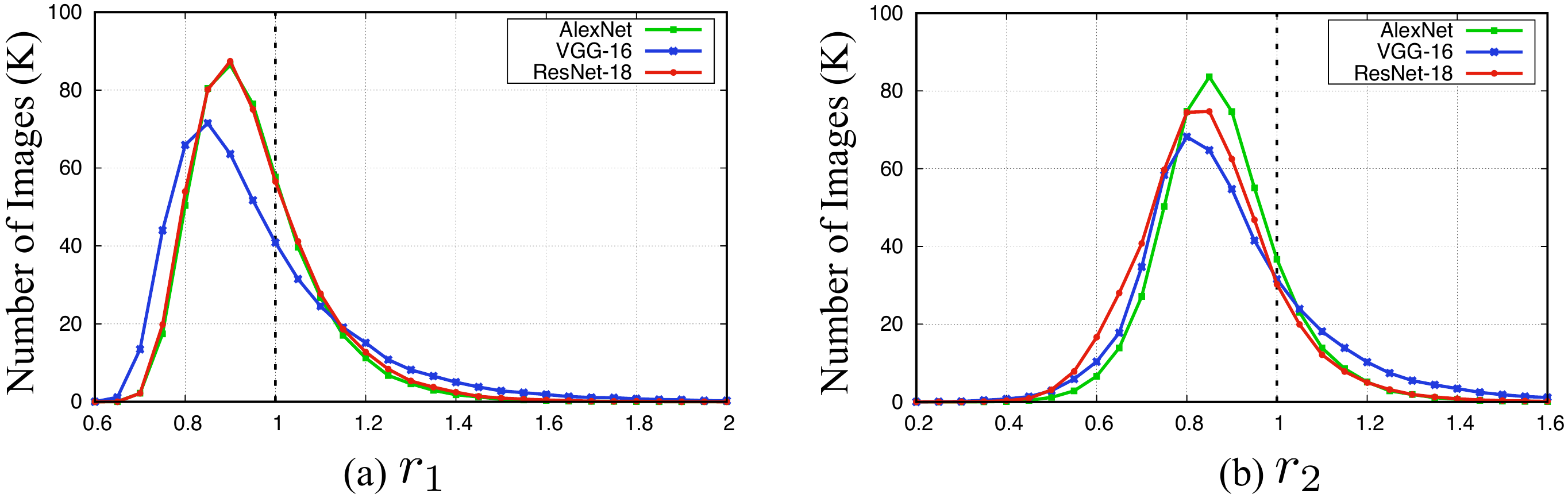}
  \caption{The distribution of adversarial images with respect to the ratio of distances after adversarial training. The average $r_1$ of AlexNet, VGG-16 and ResNet-18 are $0.98$, $0.99$ and $0.98$ for all adversarial images while the average $r_2$ of them are $0.90$, $0.92$, $0.87$ , respectively.}
  \label{fig:distance-train}
\end{figure}

We train the DNNs by minimizing an adversarial objective as
\begin{equation}
\begin{split}
L(\bm{x}, y, \theta) & = \alpha \ell(\mathbf{1}_{y}, f_{\theta}(\bm{x})) \\
& + (1 - \alpha) \ell(\mathbf{1}_{y}, f_{\theta}(G(\bm{x}, y^*))) \\
& + \beta \|\phi_{\theta}(\bm{x}) - \phi_{\theta}(G(\bm{x}, y^*)) \|_2^2,
\label{eq:adv-train}
\end{split}
\end{equation}
where $\ell$ is the cross entropy loss; $G(\bm{x}, y^*)$ is the generative process for adversarial images with target class $y^*$; and $\alpha$, $\beta$ are two balanced weights for these three loss terms.
The last loss term is the consistent loss, which is the $L_2$ distance between the representations of a pair of real and adversarial images in the last convolutional layer.
By solving the problem, we improve the interpretability of representations in terms of both positive and negative aspects, \textit{i.e.}, the learning process encourages the neurons to consistently activate when the preferred objects/parts appear, while deactivate when they disappear.

We use a variant of Fast Gradient Sign (FGS) method~\cite{goodfellow2014explaining} to generate adversarial image as
\begin{gather}
\bm{x}_0^*  = \bm{x}, \\
\bm{x}_{t}^*  = \mathrm{clip}(\bm{x}_{t-1}^*-\epsilon\cdot \mathrm{sign}\nabla_{\bm{x}} \ell(\mathbf{1}_{y^*}, f_{\theta}(\bm{x}_{t-1}^*))), \\
G(\bm{x}, y^*) = \bm{x}_{T}^*.
\end{gather}
where $\mathrm{clip}(\bm{x})$ is used to clip each dimension of $\bm{x}$ to the range of pixel values, \textit{i.e.}, [0, 255].
In experiments, we fine-tune the AlexNet, VGG-16 and ResNet-18 by Eq.~\eqref{eq:adv-train} with $\alpha=0.5$,  $\beta=0.1$ and $T=10$ on ILSVRC 2012 training set.
After training, we test the new models with the real and adversarial images in the validation set to examine the visual representations again.

We find that the interpretability is largely improved with little performance degeneration ($1\%\sim 4\%$ accuracy drop, more details on performance are provided in Appendix).
The top images in both sets with highest activations are shown in Fig.~\ref{fig:VGG16-features-adv}. The visual concepts are quite similar in them compared with Fig.~\ref{fig:VGG16-features}, indicating that the neurons are more likely to respond to true objects/parts.


Quantitative results also prove that shown in Fig.~\ref{fig:cos-train}.
The high correlation between $\bm{p}$ and $\bm{q}$ indicates that the neurons respond to true contents in images, but not the images of the corresponding classes, yielding a more interpretable and robust representation associating with the authentic input contents.

To examine the overall visual representations, we show the ratios $r_1$ and $r_2$ of distance between representations in Fig.~\ref{fig:distance-train}.
The overall representations are much closer to the representations of their original classes and farther from those of their target classes, compared with Fig.~\ref{fig:distance}.
It can be concluded that adversarial training with a consistent loss makes the overall representations more consistent between real images and adversarial images.

\section{Towards Interpretable DNNs}
\label{sec:visual}

\begin{figure*}[t]
  \centering
  \includegraphics[width=0.98\linewidth]{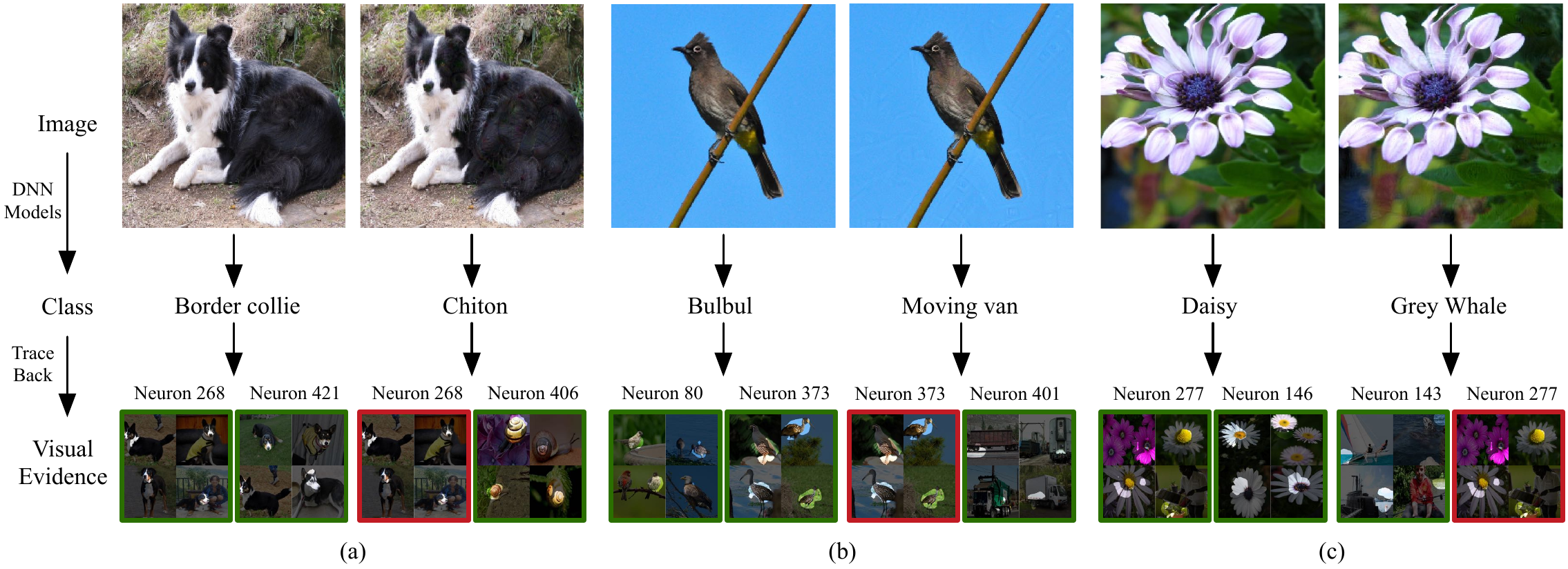}
  \caption{Demonstration for tracing the predictions back to neurons. We take VGG-16 for example with the neurons from \textit{pool5} layer. The neurons as well as their learned features can well explain the visual evidence for the predictions. The procedure also enables us to detect adversarial images manually due to the inconsistent visual evidences.}
  \label{fig:visualize}
\end{figure*}

After adversarial training, the neurons can detect visual concepts in images, but not respond to discriminative patches only. One advantage of the interpretable representations is that they provide a cue to explain why the predictions have been made.
We can trace an eventual prediction back to its representation space and figure out a set of influential neurons in the decision making process.
Here we introduce a variant of \textit{prediction difference} methods~\cite{dong2017improving,zintgraf2017visualizing} based on the hierarchical class correlation and semantic distance to retrieve a set of important neurons for a single prediction.

Formally, assume a model $f_{\theta}(\bm{x})$ is decoupled into a feature extraction module and a prediction module as $g(\phi(\bm{x}))$, where $\phi(\bm{x})$ is the feature representation of $\bm{x}$.
We aim to identify a set of influential neurons in $\phi(\bm{x})$, so we measure the relevance of each neuron $i$ with respect to the output by the prediction difference as
\begin{equation}
\mathrm{PD}(\phi(\bm{x})_i) = \| g(\phi(\bm{x})) - g(\phi(\bm{x})_{\backslash i}) \|_{\mathbf{C}}^2,
\end{equation}
where $\phi(\bm{x})_{\backslash i}$ denotes the set of all features except $i$, and the distance is measured by $\| \bm{v} \|_{\mathbf{C}}^2 = \bm{v}^T\mathbf{C}\bm{v}$.
Larger $\mathrm{PD}(\phi(\bm{x})_i)$ means a more influential neuron $i$.
For each prediction, we retrieve a set of neurons with $\mathrm{PD}$ larger than a threshold.

We show some examples in Fig.~\ref{fig:visualize}.
For each image, top-2 influential neurons with highest $\mathrm{PD}$ values are traced back and we visualize learned features of each neuron.
For real images, the visual concepts of the retrieved neurons can well explain the reasons of the current predictions.

For an adversarial image, the visual concepts of the retrieved neurons are not consistent, and some particular visual evidences are not likely to occur in the predicted class.
The neurons that detect the true visual objects in the input images will result in large prediction difference when removing them, because the predicted probability will be greatly affected by losing object/part detectors for the true concepts (\eg, the probability for the inaccurate class may be larger).
On the other hand, the neurons that detect the concepts of its target class will also lead to large prediction difference, because they contribute to such a flaw, and the prediction may be corrected by removing them.
The concepts of these neurons are contradictory in some cases.
Take Fig.~\ref{fig:visualize}~(b) for example, the adversarial image \textit{``bulbul''} is misclassified to \textit{``moving van''}, but the retrieved neurons reveal the feature \textit{``bird''}, which is weird in real \textit{``moving van''} images, so we can deduce that it is an adversarial image.
This process enables users to know when and why the model makes an error and makes them trust the model.

\section{Conclusions}
In this paper, we re-examine the internal representations of DNNs and obtain two conclusions which are contradictory to previous findings.
First, the neurons in DNNs do not truly detect objects/parts in input images, but they respond to objects/parts as recurrent discriminative patches.
Second, the visual representations are not robust distributed codes of visual concepts, which limits their effectiveness when transferred to other tasks.
To improve the interpretability of DNNs, we propose an adversarial training algorithm with a consistent loss.
Results show that the neurons learn more robust concepts and the representations are more consistent between real and adversarial images.
Moreover, human users can know how the models make predictions as well as when and why they make errors by leveraging the interpretable representations.

{\small
\bibliographystyle{ieee}
\bibliography{egbib}
}

\newpage
\noindent \begin{center} {\large  \textbf{Appendix}} \end{center}
\setcounter{section}{0}
\renewcommand\thesection{\Alph{section}}

In this appendix, we provide more details in our experiments.
In Sec.~\ref{sec:model}, we show the performance of the adopted DNN models (\ie, AlexNet, VGG-16 and ResNet-18) before and after adversarial training on the real and generated adversarial ImageNet validation sets. We also show that the models after adversarial training are more robust than original models.
In Sec.~\ref{sec:metric}, we show the effectiveness of the proposed $\mathrm{LC}$ metric to quantify the level and the consistency of the features learned by each neuron through human judgment.
In Sec.~\ref{sec:visualization}, we provide more visualization results of AlexNet and ResNet-18, supplementing Fig.~\ref{fig:VGG16-features} and Fig.~\ref{fig:VGG16-features-adv} in the main body of this paper.

\section{Model Details}
\label{sec:model}
We use the pre-trained AlexNet~\cite{krizhevsky2012imagenet}, VGG-16~\cite{simonyan2014very} from Caffe~\cite{jia2014caffe} model zoo\footnote{https://github.com/BVLC/caffe/wiki/Model-Zoo} and train the ResNet-18~\cite{he2015deep} model from scratch. The ResNet-18 network is trained with SGD solver with momentum $0.9$, weight decay $5\times10^{-5}$ and batch size 100. The learning rate starts from $0.05$ and is divided by $10$ when the error plateaus. We train it for $300$K iterations.

We denote the new models after adversarial training as ``AlexNet-Adv'', ``VGG-16-Adv'' and ``ResNet-18-Adv'', respectively. We also denote the real ImageNet validation set as ``Valid'' and the generated adversarial dataset as ``Valid-Adv''.

\begin{table}[h]
  \begin{center}
  \begin{tabular}{|c|c|c|c|}
  \hline
    Model & Dataset & top-1 & top-5 \\
    \hline
    AlexNet & Valid & 54.53 & 78.17 \\
    VGG-16 & Valid & 68.20 & 88.33 \\
    ResNet-18 & Valid & 66.38 & 87.13 \\
      \hline
    AlexNet-Adv & Valid & 49.89 & 74.28 \\
    VGG-16-Adv & Valid & 62.81 & 84.61 \\
    Res-18-Adv & Valid & 64.24 & 85.75 \\
    \hline
      \hline
    AlexNet & Valid-Adv & 0.32 & 3.41 \\
    VGG-16 & Valid-Adv & 0.32 & 10.72 \\
    ResNet-18 & Valid-Adv & 0.02 & 20.35 \\
    \hline
    AlexNet-Adv & Valid-Adv & 41.37 & 65.42 \\
    VGG-16-Adv & Valid-Adv & 45.74 & 71.20 \\
    ResNet-18-Adv & Valid-Adv & 46.42 & 71.48 \\
    \hline
  \end{tabular}
  \end{center}
  \caption{Accuracy (\%) on ImageNet}
  \label{tab:performance}
\end{table}

We report the performance of these models in Table~\ref{tab:performance} on both real validation set and adversarial validation set, and all the results are reported with only single center crop testing.
We notice that after adversarial training, the accuracies on the real validation set drop $3.89\%$, $3.72\%$ and $1.38\%$ for AlexNet, VGG-16 and ResNet-18, respectively (in the term of top-5 accuracy).
But the accuracies on the adversarial set are largely increased (\ie, $62.01\%$, $60.48\%$ and $51.13\%$ for AlexNet, VGG-16 and ResNet-18, respectively).
We argue that the adversarial training is a trade-off between models' performance and their interpretability as well as the robustness.

In order to further prove the robustness of the models after adversarial training, we conduct experiments to evaluate their performance against adversarial perturbations by Fast Gradient Sign (FGS) method~\cite{goodfellow2014explaining}.
FGS constructs adversarial images by moving the vector representation of an image along a particular direction defined by the sign of the gradient, which can be formulated as
\begin{equation}
\bm{x}^* = \mathrm{clip}(\bm{x} + \epsilon \cdot \mathrm{sign} (\nabla_{\bm{x}} \ell(\mathbf{1}_{y}, f_{\theta}(\bm{x})))),
\end{equation}
where $y$ is the ground-truth label of an image $\bm{x}$.
We test AlexNet, VGG-16, ResNet-18 and their corresponding versions after adversarial training (\ie, AlexNet-Adv, VGG-16-Adv and ResNet-18-Adv) against FGS attack with $\epsilon =1$ and $\epsilon = 5$ on ImageNet validation set.

\begin{table}[!tbh]
  \begin{center}
  \begin{tabular}{|c|c|c|c|c|}
  \hline
    \multirow{2}{*}{Model}  &  \multicolumn{2}{c|}{$\epsilon = 1$} &  \multicolumn{2}{c|}{$\epsilon = 5$} \\
    \cline{2-5}
    & top-1 & top-5 & top-1 & top5 \\
    \hline
    AlexNet & 9.04 & 32.77 & 0.41 & 5.86 \\
    AlexNet-Adv & 21.16 & 49.34 & 3.84 & 20.43 \\
    \hline
    VGG-16 & 15.13 & 39.82 & 1.95 & 14.96 \\
    VGG-16-Adv & 47.67 & 71.23 & 10.17 & 39.18 \\
    \hline
    ResNet-18 & 4.38 & 31.66 & 0.59 & 9.37 \\
    ResNet-18-Adv & 20.67 & 51.00 & 3.63 & 22.41\\
    \hline
  \end{tabular}
  \end{center}
  \caption{Accuracy (\%) on ImageNet validation set against FGS attack}
  \label{tab:fgs}
\end{table}

Table~\ref{tab:fgs} summarizes the results. The new models after adversarial training can largely improve the accuracy against attacks.
We show that our adversarial training scheme with a consistent loss can also improve the robustness of DNNs, which agrees with previous conclusions~\cite{szegedy2013intriguing,goodfellow2014explaining}.

\begin{figure}[!tbh]
  \centering
  \includegraphics[width=0.98\linewidth]{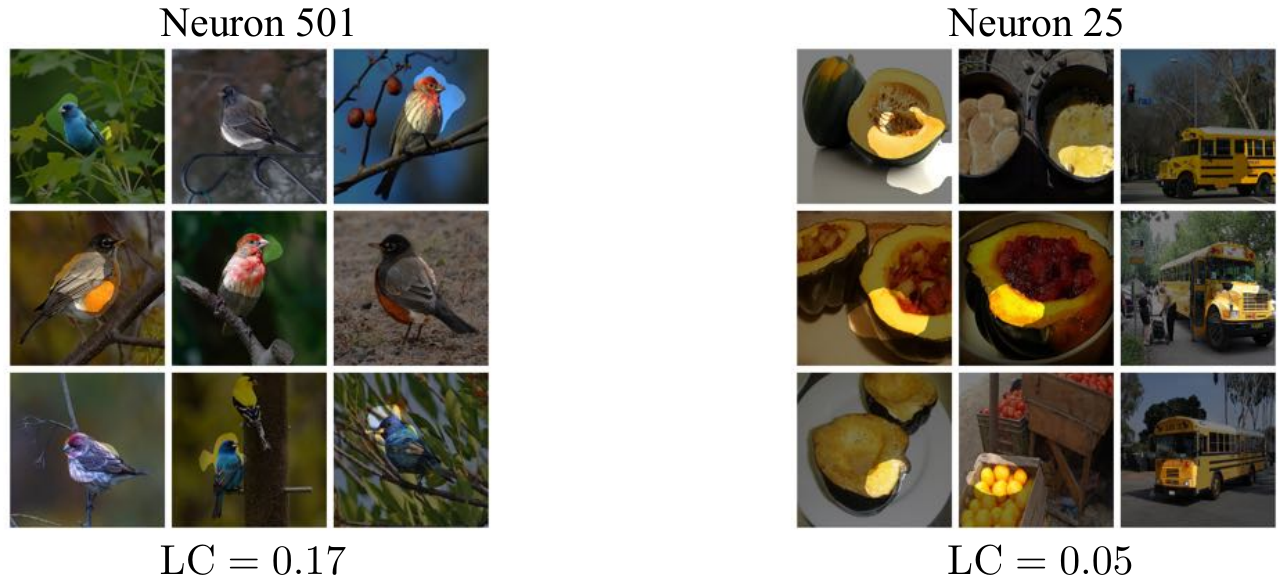}
  \caption{We show $9$ images with highest activations of two neurons for each volunteer, who was asked to answer which one represents a higher level of concepts consistently. Humans can easily identify that the left neuron detects \textit{bird} while the right neuron detects \textit{yellow objects}, so the left neuron detects a higher level concept. It agrees with the $\mathrm{LC}$ metric, which scores the left neuron $0.17$ and the right neuron $0.05$. Note that we do not show the $\mathrm{LC}$ score during human judgment.}
  \label{fig:human}
\end{figure}

\begin{figure*}[t]
  \centering
  \includegraphics[width=0.9\linewidth]{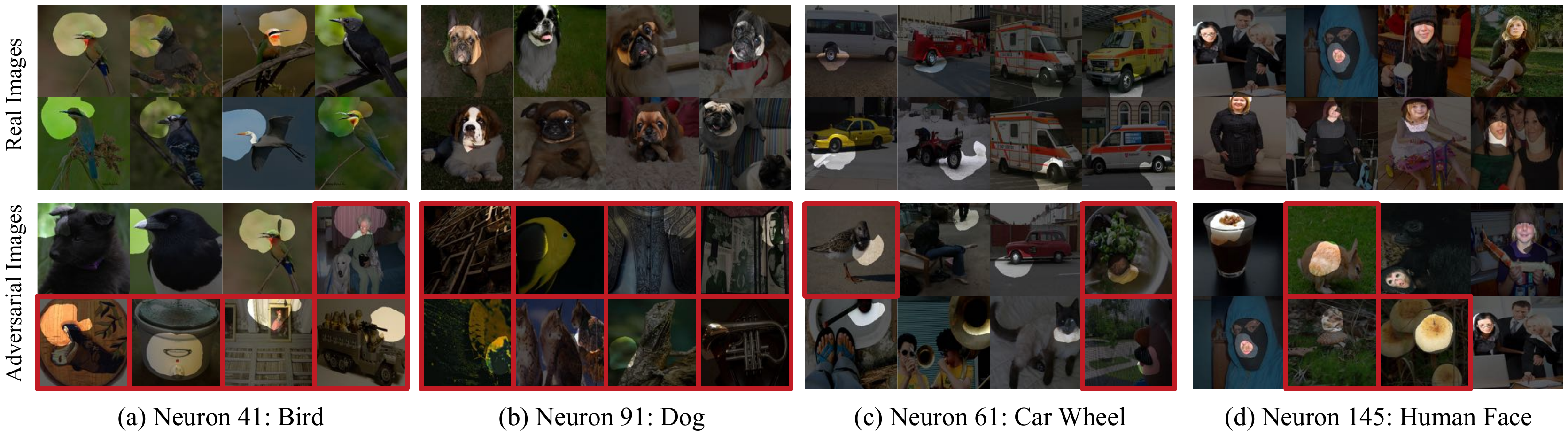}
  \caption{The real and adversarial images with highest activations for neurons in AlexNet \textit{pool5} layer. Supplementary visualization results of Fig.~\ref{fig:VGG16-features} in the main body of the paper.}
  \label{fig:alex-features}
\end{figure*}

\begin{figure*}[t]
  \centering
  \includegraphics[width=0.9\linewidth]{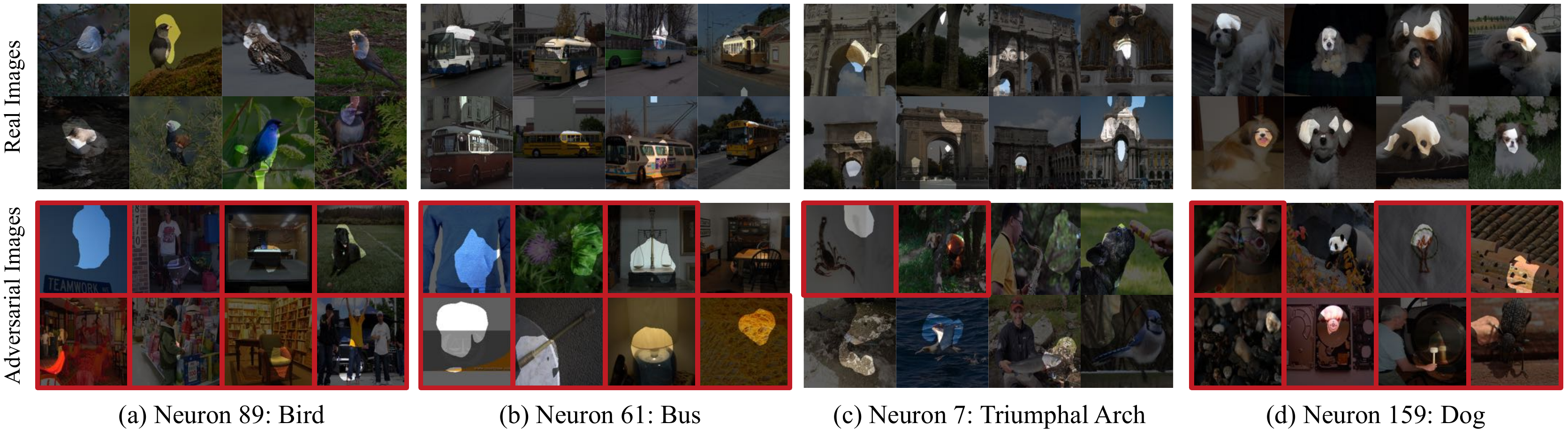}
  \caption{The real and adversarial images with highest activations for neurons in ResNet-18 \textit{conv5b} layer. Supplementary visualization results of Fig.~\ref{fig:VGG16-features} in the main body of the paper.}
  \label{fig:resnet-features}
\end{figure*}

\begin{figure*}[t]
  \centering
  \includegraphics[width=0.9\linewidth]{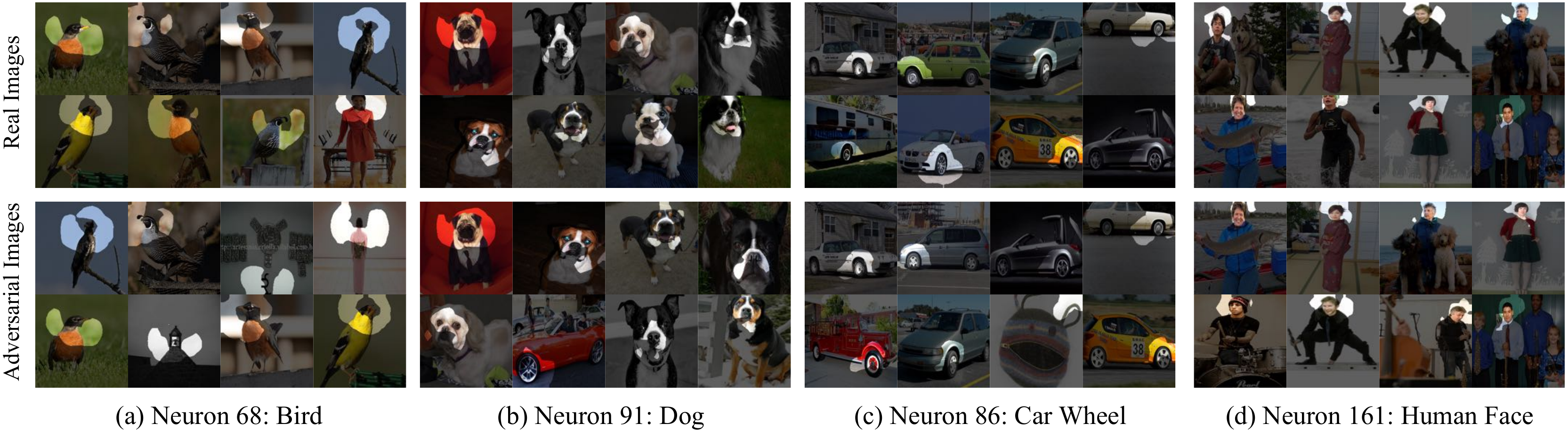}
  \caption{The real and adversarial images with highest activations for neurons in AlexNet \textit{pool5} layer after adversarial training. Supplementary visualization results of Fig.~\ref{fig:VGG16-features-adv} in the main body of the paper.}
  \label{fig:alex-features-adv}
\end{figure*}

\begin{figure*}[t]
  \centering
  \includegraphics[width=0.9\linewidth]{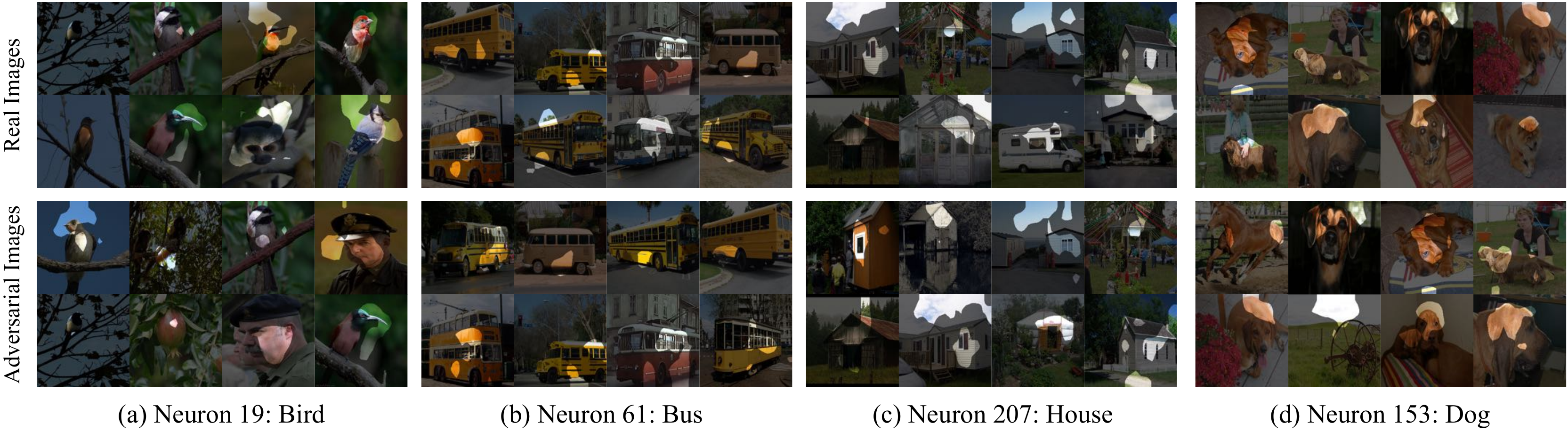}
  \caption{The real and adversarial images with highest activations for neurons in ResNet-18 \textit{conv5b} layer after adversarial training. Supplementary visualization results of Fig.~\ref{fig:VGG16-features-adv} in the main body of the paper.}
  \label{fig:resnet-features-adv}
\end{figure*}

\section{Evaluation of the $\mathbf{LC}$ Metric}
\label{sec:metric}
We evaluate the effectiveness of the proposed $\mathrm{LC}$ metric defined in Sec.~\ref{sec:quantify} by human judgment.
We choose VGG-16 model for evaluation.
In experiments, we ask $5$ volunteers to judge the level and the consistency of the featured learned by neurons.
Each volunteer was asked to compare two neurons, each one of which is represented by $9$ images with highest activations. Each image is also shown with highlighted regions found by discrepancy map~\cite{zhou2014object}.
We random select $2000$ pairs of neurons in VGG-16 and show the corresponding images for each volunteer, who was asked to answer which one of the two neurons consistently represents a higher level of concepts, as shown in Fig.~\ref{fig:human}.
The volunteers were not aware of the definition of the metric as well as our purpose.

We also compare the $\mathrm{LC}$ metric of these pairs of neurons, \ie, a higher score indicates a higher level concept. We measure the agreements of the results found by our methods and human judgments. The consistency between our metric and each volunteer is $83.65\%$, $85.75\%$, $86.6\%$, $88.6\%$ and $86.95\%$. It proves that our proposed metric is effective to measure the level and the consistency of each neuron's features.

\section{Visualization Results}
\label{sec:visualization}
We show supplementary visualization results in Fig.~\ref{fig:alex-features} and Fig.~\ref{fig:resnet-features}. The dummy object/part detectors also emerge in AlexNet and ResNet-18, which is the same as VGG-16 , as we show in Fig.~\ref{fig:VGG16-features} in the main body of the paper.

After adversarial training, the neurons learn to resist adversarial perturbations and respond to true objects/parts. We also show the visualization results of AlexNet and ResNet-18 after adversarial training in Fig.~\ref{fig:alex-features-adv} and Fig.~\ref{fig:resnet-features-adv}. The results are consistent with those in Fig.~\ref{fig:VGG16-features-adv} in the main body of the paper.

\end{document}